\documentclass[letterpaper, 10 pt, conference]{ieeeconf}
\IEEEoverridecommandlockouts
\usepackage{cite}
\usepackage{amsmath,amssymb,amsfonts}
\usepackage{algorithmic}
\usepackage{graphicx}
\usepackage{textcomp}
\usepackage{xcolor}
\usepackage{makecell}
\usepackage{mars}
\usepackage[obeyspaces]{url}
\usepackage{hyperref}  
\PassOptionsToPackage{hyphens}{url}

\def\BibTeX{{\rm B\kern-.05em{\sc i\kern-.025em b}\kern-.08em
    T\kern-.1667em\lower.7ex\hbox{E}\kern-.125emX}}

\title{\LARGE \bf Cluster on Wheels}
\author{Yuanyuan Yang$^{1}$, Delin Feng$^{1}$ and S\"oren Schwertfeger$^{1}$
\thanks{$^{1}$All authors are with the School of Information Science and Technology, 
	ShanghaiTech University, China.
	{\tt\small [yangyy2, fengdl, soerensch]@shanghaitech.edu.cn}}%
}
\begin{document}
	
%
%


\marsPublishedIn{Published with:} 	

\marsVenue{IEEE International Conference for Advancement in Technology (ICONAT) 2022}

\marsYear{2022}

\marsPlainAutors{Yuanyuan Yang, Delin Feng and S\"oren Schwertfeger}


\marsMakeCitation{Cluster on Wheels}{IEEE Press}

\marsDOI{10.1109/ICONAT53423.2022.9725992}

\marsIEEE{}


\makeMARStitle

%
%

%

\maketitle

\begin{abstract}
This paper presents a very compact 16-node cluster that is the core of a future robot for collecting and storing massive amounts of sensor data for research on Simultaneous Localization and Mapping (SLAM). To the best of our knowledge, this is the first time that such a cluster is used in robotics. We first present the requirements and different options for computing of such a robot and then show the hardware and software of our solution in detail. The cluster consists of 16 nodes of AMD Ryzen 7 5700U CPUs with a total of 128 cores. As a system that is to be used on a Clearpath Husky robot, it is very small in size, can be operated from battery power and has all required power and networking components integrated. Stress tests on the completed cluster show that it performs well.
\end{abstract}
{\bf\emph{ Index Terms}-\ Robotics, Computing, Cluster, Sensor Data\rm}

\section{Introduction}

Localization and Mapping, often used together in a  Simultaneous Localization and Mapping (SLAM) system, are essential components for many mobile autonomous robotic systems \cite{cadena2016past}. 
For the development and testing of SLAM algorithms it is essential to use datasets. Public datasets enable researchers to objectively compare their approaches. Recent published datasets include \cite{shi2020we} and \cite{wang2020tartanair}, while \cite{chen2020advanced} includes a survey of various robotic datasets for SLAM.

SLAM can be done using different sensors, for example 2D and 3D LiDARs, RGB and RGB-D depth cameras, event cameras, ultrasound or radar. Often sensor-fusion is used with two or more sensors, like stereo vision, non-overlapping sensor setups or the inclusion of additional data from wheel odometry, Inertial Measurement Units (IMUs), GPS or WiFi localization. This leads to a fragmentation of the benchmarking, since datasets typically provide only a subset of said sensor configurations and researchers are thus often forced to use their own datasets for benchmarking, since public datasets do not fulfill the data requirements of their SLAM system. 

Some of the aforementioned sensors can provide massive amounts of data. This is especially true for RGB, RGB-D and event cameras, which often could provide more than 600MBps (USB 3.0 speed). 
For example the cameras used in \cite{chen2020advanced} (GS3-U3-51S5C-C) can provide 2,448 x 2,048 pixel at up to 75Hz, which corresponds to 1.05GBps of 8-bit RGB data after debayering. Consequently, datasets typically reduce resolution or frame rate to avoid huge datasets. But it is conceivable that future robotic systems may utilize specialized data processing processors that can handle such huge amounts of data - if it turns out to be advantageous for SLAM or another application (e.g. computer vision). 

It is thus desirable to collect datasets with as high properties as possible (e.g. frame rate and resolution) and with as many different sensors in different configurations simultaneously, to enable the development and comparison of present and future SLAM algorithms. Building robots to collect such datasets is a challenging task. In \cite{chen2020advanced} we presented a system that used 9 cameras (stereo: front, left, right, up; monocular: back), two Velodyne HDL-32E 3D LiDARs and an IMU. The robot used a single Intel Core i7-6770k CPU and additional USB 3.0 cards to connect the cameras, but was only able to collect the full resolution camera data at 10 Hz due to CPU limitations.

\begin{figure}[tp]
	\centerline{\includegraphics[width=1.0\linewidth]{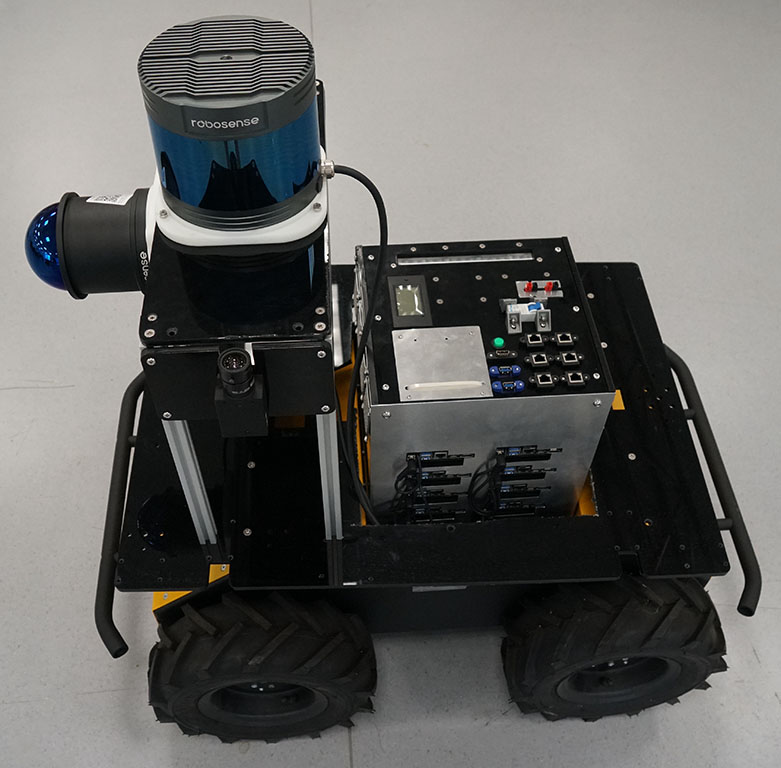}}
	\caption{The cluster (the silver and black box in the center of the robot) mounted on an early mockup of the ShanghaiTech Mapping Robot II.}
	\label{fig:mapping_robot_II}
\end{figure}

We are now in the process of building the ShanghaiTech Mapping Robot II, which should feature many more sensors (e.g. 11 cameras: additionally stereo back and monocular down, all at 60 frames per second; three event cameras; stereo infrared cameras; an omni-directional camera; 5 RGB-D cameras; 4 high-resolution 3D LiDARs, as well as sonar \& radar sensors). Fig. \ref{fig:mapping_robot_II} shows a mockup of that robot with just a few sensors and our cluster at its center. The computational needs for this robot are massive: there should be enough high-bandwidth I/O ports (USB 3.0 and Gigabit Ethernet); enough CPU processing power to compress images to high-quality JPEG images and enough bandwidth and size for file storage on SSDs. We did not opt to utilize specialized hardware for video capture and encoding like \footnote{\url{http://www.magewell.com/products/eco-capture-hdmi-4k-m2} or \\
\url{https://buy.advantech.com/Boards-Cards/Video-Cards/Indus_Video_Sulotion.products.htm}
}, because we prefer to keep the system general and capable of collecting data from all kind of current and future sensors.

Computation for robotics often includes special hardware like FPGAs \cite{wan2021survey} or GPUs \cite{qiu2009gpu}, but there is scarce literature on the topic of on-board high-performance CPU computing for robotics. In \cite{camargo2018towards} a theoretical foundation on using a cluster for robotics is presented and experiments are conducted with a toy experiment consisting of four Raspberry PI 3 boards. The authors further explore this topic in \cite{camargo2019archade} where ubiquitous supercomputing for robotics is explored, that includes distributed computing over several robots and the cloud. 
Computation offloading \cite{kumar2013survey} and cloud computing for robotic are hot topics \cite{kehoe2015survey}, but not applicable for the high I/O demand scenario presented here. 

In \cite{ribeiro2015robotic} another small system of 5 ODROID-x2 nodes for computation on a robotic blimp is presented. 

Multi-processor approaches to specific robotic systems have been proposed early on, for example 1992 on the problem of inverse dynamics \cite{fijany1992parallel}, but such papers are often theoretic/ algorithm oriented, don't present concrete systems and usually assume a unified memory and not a cluster system. 

Willow Garage's PR2 robot had two high-performance computers (2 $\times$ Intel i7 Xeon quad-core processors, 24 GB RAM each), of which the second is network booting from the first \footnote{\url{https://www.clearpathrobotics.com/assets/downloads/pr2/pr2_manual_r321.pdf}} - in so far one could speak of a very small cluster. The Robot Operating System (ROS) \cite{quigley2009ros} developed by Willow Garage natively  supports multi-node messaging, a fact that is heavily exploited in the software architecture of this project.

An article on {\it ROS 2 and Kubernetes Basics} \footnote{\url{https://ubuntu.com/blog/exploring-ros-2-with-kubernetes}} explores how to distribute ROS 2 compute tasks across multiple machines with Kubernetes, in the example using three nodes. Actually, we are considering following this software design using ROS 2 and Kubernetes in the future.  

Multi-node computing is more prevalent in autonomous driving, presumably because of the availability of more power and space. E.g. \cite{wei2013towards} report on a multi-node system and NVIDIA is supporting distributed multi-node autonomous vehicle AI training with the NVIDIA DGX Robotic Drive systems \footnote{\url{https://developer.nvidia.com/blog/validating-distributed-multi-node-av-ai-training-with-dgx-systems-on-openshift-with-dxc-robotic-drive/}}. 

This paper describes the requirements, options and final solution for the computation needs of the ShanghaiTech Mapping Robot II. We believe that our approach of using a cluster for robotics with more than just a few nodes has never been used or reported in literature. Section \ref{sec:computing} presents the computing requirements for the mapping robot and different options. Section \ref{sec:hardware} describes the hardware of the 16-node cluster that we designed for our wheeled data collection robot while Section \ref{sec:software} presents its software stack. 
Section \ref{sec:conclusions} concludes this paper.

%
%
%
%
%

\section{Computing Requirements and Options}
\label{sec:computing}

The requirements for the computing solution for the ShanghaiTech Mapping Robot II are dictated by the robot and the planned sensor suite, but should also leave room for future upgrades and changes. Physically the system has to fit in the payload bay of the Clearpath Husky robot, with a width and length of 380 mm and 290 mm, respectively. The height should be moderate, because several 360 degree field-of-view sensors (omni-camera, 3D LiDAR) have to be placed above the computing solution. The power consumption should not be too high for battery powered operation, not much higher than 1kW. The main computation requirements are:
\begin{enumerate}
    \item I/O: receive data at full USB 3.0 bandwidth of 20 devices (11 RGB cameras, 1 omni camera, 3 event cameras, 5 RGB-D cameras) and several other high-bandwidth Gigabit Ethernet devices (4 3D LiDARs, 2 infrared cameras)
	\item CPU power: ability to debayer and JPEG compress (quality 90) at least 11 streams of 5MP @ 60Hz frame rate, while handling the I/O of the other data.
	\item Storage: have enough bandwidth and size to store the data (e.g. 9 full bandwidth USB 3.0 devices (9 x 600 MBps) + compressed camera (11 x 150 MBps) + other devices (6 x 100MBps) $\approx$ 7.6 GBps $\approx$ 27 TB per hour)
\end{enumerate}

Given the available battery, mapping runs can be limited to one hour, requiring about 30TB of disk space at full specs. While a size of 30 TB for a robot dataset may seem daunting, it can be stored in an Amazon S3 Glacier instance for about 120 USD per month. A download with a Gigabit Internet connection can be achieved in less than 3 days, incurring a cost of about 80 USD from Amazon.

\subsection{Computing Options}

\begin{table*}[htbp]
	\caption{Computer Options Comparison}
	\begin{center}
		\begin{tabular}{|c|c|c|c|c|c|c|}
			\hline
			\textbf{Motherboard}&\textbf{CPU}&\textbf{CPU TDP}&\textbf{Cores / Threads}&\textbf{Size} &\textbf{I/O} \\
			\hline
			\multicolumn{6}{c}{One-Node option ( 1 $ \times $ )} \\ \hline 
			
			ASROCK TRX40 Creator& \makecell[c]{AMD Ryzen \\Threadripper 3970X} & 280W &32  / 64 &305 $ \times $ 244 mm
			&\makecell[l]{
				4 USB 3.2 Gen2 \\
				8 USB 3.2 Gen1 \\
				2 Gigabit+ Ethernet \\
				3 M.2 \\
				8 SATA3 \\ \hline
				3 x PCIe 4-port USB 3.0 cards \\
				1 x PCIe 4-port Ethernet card
			}\\
			\hline
			\multicolumn{6}{c}{Four-Node Options ( 4 $ \times $ )} \\ \hline 
			
			ASROCK Z590 Taichi & Intel Core i9-11900 & \makecell[c]{65W \\ (260W)} & \makecell[c]{8 / 16 \\ ( 32 / 64 )} &300 $ \times $ 240 mm
			&\makecell[l]{2 (8) Thunderbolt 4 USB \\
				6 (24) USB 3.2\\
				2 (4)$^{\mathrm{a}}$ Gigabit+ Ethernet \\
				3 (12) M.2\\
				8 (32) SATA3\\ \hline
				1 (4) 2-port (8) PCIe Ethernet card
			}\\
			\hline
			ZOTAC ZBOX MI574 & Intel Core i7-9700& \makecell[c]{65W \\ (260W)} & \makecell[c]{8 / 16 \\ ( 32 / 64 )} &185 $ \times $ 185 mm
			&\makecell[l]{1 (4) USB 3.1 Type-C\\
				6 (24) USB3.0\\
				2 (4)$^{\mathrm{a}}$ Gigabit+ Ethernet \\
				2 (8) M.2 \\
				1 (4) SATA33\\ \hline
				1 (4) 2-port (8) PCIe Ethernet card}\\
			\hline
			\multicolumn{6}{c}{16-Node Options ( 16 $ \times $ )} \\ \hline 
			
			\makecell[c]{ASUS Mini PN51 \\ (used in this project)} & AMD Ryzen7 5700U& \makecell[c]{15W \\ (240W)} & \makecell[c]{8 / 16 \\ ( 128 / 256 )} &115 $ \times $ 115 mm
			&\makecell[l]{2 (32) USB 3.2 Gen2 Type C\\
				3 (48) USB 3.1 Gen1 \\
				1 (0)$^{\mathrm{a}}$ Gigabit Ethernet \\
				1 (16) M.2  \\
				1 (16) SATA3\\ \hline
				1 (16) 1-port (16) M.2 Ethernet card}\\
			\hline
			ASRock 4X4-4800U& AMD Ryzen 7 4800U& \makecell[c]{15-25W \\ (240-400W)} & \makecell[c]{8 / 16 \\ ( 128 / 256 )} & \makecell[c]{4 $  \times $ 4 \\ (104 $ \times $ 102 mm)}
			&\makecell[l]{
				4 (64) USB 3.2 Gen2\\
				4 (64) USB 2.0\\
				2 (16)$^{\mathrm{a}}$ Gigabit Ethernet \\
				1 (16) M.2 \\
				1 (16) SATA3}\\
			\hline
			ASRock 4X4-V2000M& \makecell[c]{AMD Ryzen\\ Embedded V2718} & \makecell[c]{10-25W \\ (160-400W)} & \makecell[c]{8 / 16 \\ ( 128 / 256 )} &  \makecell[c]{4 $  \times $ 4 \\ (104 $ \times $ 102 mm)}
			&\makecell[l]{2 (32) USB 3.2 Gen2\\
				4 (64) USB 2.0\\
				2 (16)$^{\mathrm{a}}$ Gigabit Ethernet \\
				1 (16) M.2 \\
				1 (16) SATA3}\\
			\hline
			NUC11 PAHi7& Intel Core i7-1165G7& \makecell[c]{12-28W \\ (192-448W)} & \makecell[c]{4 / 8 \\ ( 64 / 128 )} &117$ \times $112 mm
			&\makecell[l]{2 (32) Thunderbolt 4 \\
				3 (48) USB 3.1 \\
				1 (0)$^{\mathrm{a}}$ Gigabit Ethernet \\
				1 (16) M.2 2280\\
				2 (32) SATA3 }\\
			\hline
			GIGABYTE/BSI7-1165G7& Intel Core i7-1165G7& \makecell[c]{12-28W \\ (192-448W)} & \makecell[c]{4 / 8 \\ ( 64 / 128 )}  &196 $ \times $ 140 mm
			&\makecell[l]{1 (16) Thunderbolt 4 \\
				6 (96) USB 3.2 \\
				2 (16)$^{\mathrm{a}}$ Gigabit Ethernet \\
				2 (32) M.2\\
				1 (16) SATA3}\\
			\hline
			\multicolumn{6}{l}{$^{\mathrm{a}}$In a cluster setup one Ethernet port per node is used by the cluster and thus not available for connecting sensors.}
		\end{tabular}
		\label{tab:Comparison}
	\end{center}
\end{table*}

There were several options under consideration to fulfill the above computation requirements:
 A) a single high-power CPU node with multiple USB 3.0 I/O cards and a big RAID of SSD disks; B) four workstation PCs, each with USB 3.0 I/O cards and a RAID of SSD disks; C) a cluster of mini PC boards, utilizing their on-board USB controllers, each with a single SSD. For 32TB of storage Option A) would utilize a RAID of 8 4TB SSDs, Option B) would use two 4TB SSDs per node while Option C) can use one 2TB SSD per node. Table \ref{tab:Comparison} shows some specifications of the different computing nodes and some of the theoretical cluster specifications. It should be noted that the TDP (thermal design power) listed is just for the CPU, while other components may also draw significant power (e.g. board, main memory, SSD, fans), potentially more than doubling the TDP value for the consumption of the whole system.

In order to compare the computation performance of the CPUs we utilized public results from the well-known cross-platform benchmark \textit{Geekbench 5}. Benchmarking is a complex problem and all kinds of factors (e.g. cooling, configuration, operating system) play a role here, but nevertheless the results published on their website allow for a rough comparison of the processor speeds. The software measures  single-threaded and multi-threaded performance with multiple algorithms and analyzes and scores the operating performance of the processor in an all-round way. The benchmark results are shown in Table \ref{tab:benchmark}.

\begin{table}[htbp]
	\caption{Geekbench 5 Scores}
	\begin{center}
		\begin{tabular}{|c|c|c|}
			\hline
			\textbf{CPU}& \textbf{Single-Core} &\textbf{Multi-Core (Cluster)} \\
			\hline 
			AMD Ryzen Threadripper 3970X & 1,264 & 24,183 \\
			\hline
			Intel Core i9-11900 & 1,549 & 6,539 (26,156) \\
			\hline
			Intel Core i7-9700& 1,198 & 6,397 (25,588)\\
			\hline
			AMD Ryzen 7 5700U$^{\mathrm{a}}$& 1,044 & 5,526 (88,416)\\
			\hline
			AMD Ryzen 7 4800U& 1,029 & 5,905 (94,408) \\
			\hline
			AMD Ryzen Embedded V2718& 1,165 & 6,856 (109,696) \\
			\hline\
			Intel Core i7-1165G7& 1,398 & 4,573 (73,168) \\
			\hline
			\multicolumn{3}{l}{$^{\mathrm{a}}$Used in this project.}
		\end{tabular}
		\label{tab:benchmark}
	\end{center}
\end{table}

\subsection{Computing Options Tests \& Discussion}

In the following the advantages and disadvantages of the different options are discussed and our final selection is justified.

\subsubsection{AMD Ryzen Threadripper 3970X}
Using a single, very powerful computer was our initial approach, as it allows for a fairly quick and easy integration into the robot (even though water-cooling may be advised) and also eases the use of the robot, having to operate only one computer. An initial test in our lab with two 4-port USB 3.1 PCIe 4-lane cards (StarTech PEXUSB3S44V) relieved that using more than 8 GS3-U3-51S5C-C cameras results in frame drops and disconnected cameras, using the capture tool provided by the camera manufacturer or ROS, regardless of the combinations of USB ports used from the motherboard or the PCIe cards. This is of course unacceptable for use in our robot.
This option would also severely limit a possible future extension of the requirements, as the I/O and storage system are maxed out in this setup already. 

\subsubsection{Four-Node Options}
Our lab already has experience with such a setup, as we operate a big box with four such boards next to each other to easily add computation and sensors to cars for research on smart cars and autonomous driving. For the mapping robot four nodes could be mounted vertically, on top of each other. So the size and also the power requirements outlined earlier are met. The I/O requirements can also be met (24 USB3.2 ports are available over the 4 systems) when adding a PCIe card for additional Ethernet ports. 

A considerable advantage of these boards (as well as the Threadripper above) is that they can be easily extended using the PCIe slots, for example with dedicated cards for some specialized sensors or with FPGAs or GPUs to enhance computation or encoding performance. 

A problem for this option is the low overall CPU performance. As mentioned in the requirements, the data from 11 cameras should be debayered and JPEG compressed @ 5MP @ 60 Hz. Besides handling all the other I/O, each node would have to process three such streams. We do not have such CPUs in our lab, but in the boxes mentioned about we utilize 6-core i7 processors. These are able to handle two such streams with the standard ROS drivers - with almost full CPU load. We are working on writing optimized storage nodes, that, for example, also should speed things up by utilizing libjpeg-turbo\footnote{This library claims to be 2-6 times faster in JPEG encoding, compared to libjpeg: \url{https://libjpeg-turbo.org/}; While ROS is utilizing libjpeg-turbo through OpenCV, its actual turbo (SIMD) is deactivated at the time of the writing of this article: \url{https://github.com/opencv/opencv/issues/12115}. }, but having three such streams on the 8-core Intel CPUs is pushing the boundaries, especially since the four PCs also have to handle all the other I/O. The 16-Node option offers more performance and also offers CPU and I/O performance for future extensions.

The ZOTAC ZBOX Mi574 is smaller than the ASROCK Z590 motherboard, but has a slower CPU and offers fewer extension ports.

%

\subsubsection{16-Node Options}
The requirements state that 11 RGB cameras need to be compressed, and additional 9 full-bandwidth USB 3.0 devices plus 6 high-bandwidth Ethernet sensors need to be handled. With 16 nodes we can assign 1 (or 0) RGB cameras with JPEG compression to each node, plus either another USB 3.0 device or high-bandwidth Ethernet sensor. Following this setup already leaves capacity for 6 more devices, while our tests showed that the node we selected can actually comfortably handle an additional full-bandwidth device (with or without JPEG compression). 

The benchmark also shows that the 16-Node Options have far superior overall multi-core performance. This allows for alternative use cases, for example for high-performance computations for live SLAM from a number of high-bandwidth sensors. This would permit for mapping runs that do not store all raw data on the disks (even though, depending on the actual CPU load, battery changes every 1 or 2 hours will be required).

For all options it is quite difficult to get the data off the cluster. While a download of assumed 30TB over the cluster network will take about 2.8 days, attaching an HDD (which are available in sizes of up to 18TB as of 2021) to each of the 16 nodes will reduce the time to about 130 minutes. 

\subsubsection{ASRock 4$ \times $4}
\textit{ASRock 4$ \times $4-4800U} is a motherboard released in 2020. It is equipped with a \textit{Ryzen 7 4800U} processor. The base frequency is 1.8GHz, the acceleration frequency is 4.2GHz. One of its advantages is that it has 2 wired network ports. Additionally, using a sample board we purchased, we found that it is possible to turn the CPU fan by 180 degree. The plan was to do this for all boards, such that the power connection and ethernet ports are all inside, while still blowing the hot air outside. This would have allowed a cleaner power-connection setup compared to the solution presented below.

The \textit{ASRock 4$ \times $4 motherboard} series also has a board named \textit{ASRock 4$ \times $4-V2000M} which is very similar to \textit{4$ \times $4-4800U}, except for the processor. \textit{4$ \times $4-V2000M} is an industrial grade product, its production cycle (7-15 years) is longer than \textit{4$ \times $4-4800U} (1-2 years).

Both these CPUs are faster (and more expensive) than the AMD Ryzen 7 5700U that we ultimately chose for this project, such that, together with the other advantages mentioned above, these were our clear favorite options. But our vendors were unable to provide the 16 units needed for the cluster, such that we had to give up on choosing them.

\subsubsection{NUC11 PAHi7}
\textit{NUC11 PAHi7} was released in 2021. It is equipped with an \textit{Intel Core i7-1165G7} processor. The CPU frequency is 2.8 GHz, and the highest turbo frequency reaches 4.7 GHz. It has two Thunderbolt interfaces, but only one Ethernet port. The CPUs multi-thread performance is significantly lower than the Ryzen 7 boards (manly due to having only half the number of cores).

Another mini computer is also equipped with the \textit{Intel Core i7-1165G7} processor: \textit{GIGABYTE/BSI7-1165G7}. It has two M.2 interfaces for SSD and one Thunderbolt interfaces. But the size is quite large compared with the other models.

\subsubsection{ASUS  Mini PN51}

The \textit{ASUS Mini PN51} is a mini computer released in 2021. It is equipped with a \textit{Ryzen 7 5700U} processor with a 7nm manufacturing process. The base and turbo frequencies are 1.8GHz and 4.3GHz, and the CPU power consumption is only 15W. A disadvantage of this board is, that it only has one Ethernet port, so we replaced the WiFi M2 card (which we do not need) with an Ethernet card. This works but makes the space inside the cluster more crowded. 

\begin{figure}[tbp]
	\centerline{\includegraphics[width=0.9\linewidth]{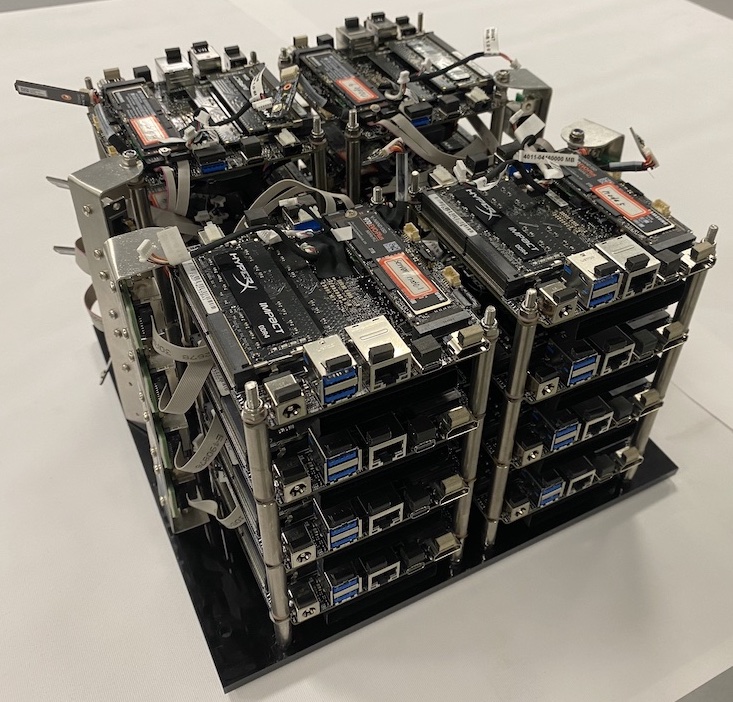}}
	\caption{The 16 nodes of the cluster, organized in four columns of four boards each. }
	\label{fig:boards}
\end{figure}

Overall we selected this as the best available option: 16  \textit{Ryzen 7 5700U} CPUs, with 32GB RAM each (512GB total), a fast 2TB M.2 SSD (Samsung 970EVO Plus 2TB (each 1,700 MBps sustained sequential write performance\footnote{\url{https://www.tomshardware.com/reviews/samsung-970-evo-plus-ssd,5608.html}}, almost enough to store 3 full-bandwidth USB 3.0 data streams (3$\times$600MBps)) and the M.2 Ethernet card. 

For this option we actually had to purchase the complete mini PCs, from which we then extracted the board and upgraded its RAM and SSD. The cost was about 750 USD per PC and 700 USD for the upgrades, such that the total price of the cluster, including the costs for the other components (DC/DC etc.), is about 25,000 USD.

\section{Cluster Hardware}
\label{sec:hardware}

The 16 nodes of our cluster are depicted in Figure \ref{fig:boards}. They are organized in four columns of four nodes each. The back-side of the board with the power connector, the LAN port, two USB 3.1 ports, one USB-C 3.2 port, an HDMI connector as well as the fan are pointing to the outside, while one USB 3.1 and one USB-C 3.2 port are located inside the cluster housing. These could be made available via extension cables, if needed.

This setup is not sufficient to run the cluster. Additionally needed is a stable and protective housing, that also provides proper cooling. Furthermore is a power distribution system needed as well as networking. Similar to the system described in \cite{chen2020advanced}, the mapping robot will also require hardware sensor synchronization. We will again utilize an Asus Tinker Board for this, as is provides real-time signaling as well as sufficient CPU speed to run the needed ROS nodes. This board will also installed in the housing (on the top plate) and can be seen as node 17, even though it is running its own operating system and not using netboot. 

\subsubsection{Mechanical and Airflow Design}
\begin{figure}[tbp]
	\centerline{\includegraphics[width=0.8\linewidth]{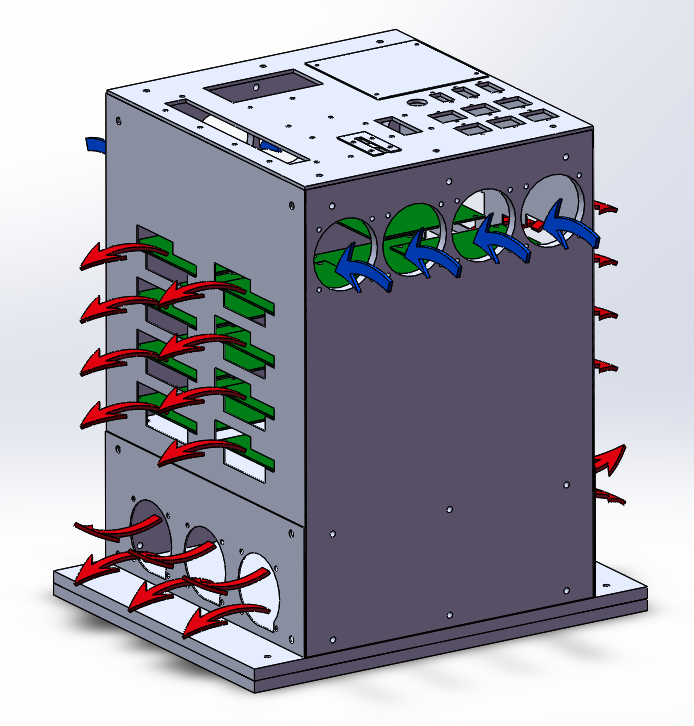}}
	\caption{The CAD design of the cluster housing with the air-flow plan. Outside air is sucked into the housing by two times four 6-cm fans on the top and warm air is pushed out by each node and a total of four 6cm fans at the bottom, for cooling the DC/ DC converters and the network switch.}
	\label{fig:CAD_design}
\end{figure}

Figure \ref{fig:CAD_design} shows the airflow design for the cluster. Eight 6cm fans are blowing outside air into the housing, while each node has a CPU fan blowing air outside. Additionally, four 6cm fans blow air out of the bottom compartment of the housing for cooling the DC/DC converters and the 24-port network switch.

The housing itself is made from 7mm thick black acrylic glass (PMMA: Polymethyl Methacrylate) that is laser-cut from the CAD design. Six 20$\times$20mm aluminum profile bars connect the walls with the bottom, middle and top plate. The front covers are CNC'ed from 1mm aluminum sheets. Fig. \ref{fig:CAD} shows the housing design of the cluster.

\subsubsection{Power Distribution}

The top plate has two pairs of Anderson Powerpole connectors to connect batteries or AC/ DC power supplies. Each connector goes to a Diode (85HF10) - this is to prevent batteries charging each other. The idea behind the two power connectors is, that the user can temporarily connect a second battery while changing the main robot battery, thus avoiding a shutdown of the cluster. Cabling is done with $15 mm^2$ copper cabling. Both inputs are connected to the circuit breaker (C25) which may trip for currents higher than 125 A. The power monitor attached to the top plate uses a big FL shunt that is connected to the circuit. Two wires then connect the top plate with the bottom compartment. Fig. \ref{fig:CAD} shows the wiring in the bottom compartment. 

The ground is distributed to all consumers via one big 24-port distribution block. The battery power is directly fed to the network switch and to the five DC/ DC converters. For powering the nodes four DC/ DC converters (input: 22V - 40V; output: 19V 20A max)  are used, each powering four nodes. A 10,000 uF capacitor is added to each of the four output circuits (without it nodes on the same DC/ DC crash if one node is plugged in). An additional DC/ DC with 12V output is powering the housing fans. 

\subsubsection{Middle Compartment}
The 16 nodes are mounted in four columns on a plate with a big hole (Fig. \ref{fig:boards}). Each column is held by four rods with M4 threading and metal spacers (3cm at the bottom, 4cm between nodes). Plastic washers protect the boards from the spacers (which are a little too thick and may touch components on the board). The external RJ-45 Ethernet sockets are mounted on four thin aluminum sheets that are screwed in the middle plate and connected to a rod at the top. 

The original mini PCs used a small PCB for two indication LEDs (power and HDD usage) and the button, connected with a short cable. We purchased 55cm cables with the appropriate connectors on both ends (SH1mm 6P) and mounted these small PCBs on the top board. We organized the four cables for each column of nodes with cable mesh.

\subsubsection{Assembly}

\begin{figure}[tbp]
	\centerline{
		\includegraphics[width=0.8\linewidth]{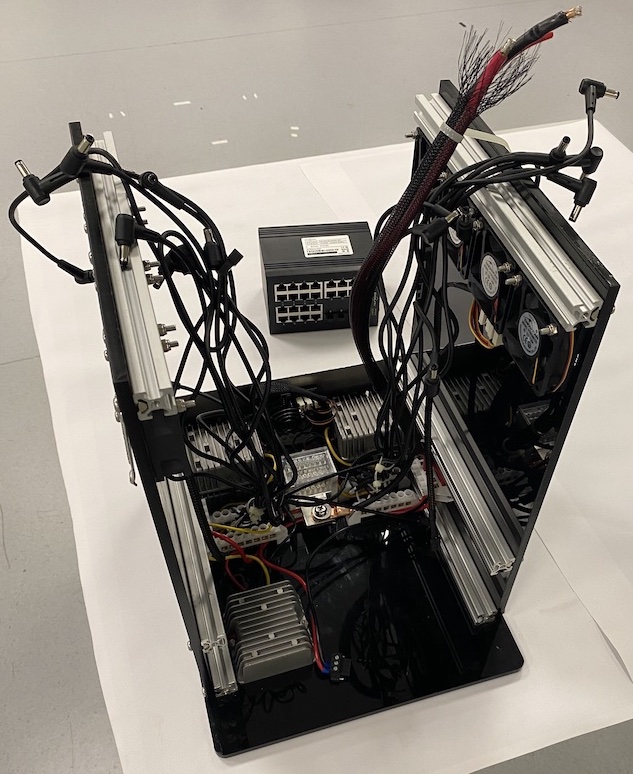}}
	\caption{The cluster housing. The bottom compartment is for power and networking, where we can see the four 19V DC/ DC converters powering the nodes (in the back in two columns of two devices), one 12V DC/ DC for the fans (front) and the 24-port switch that will also go to the bottom. The 16 power plugs for the nodes are connected to one of the four white power distributors and the big central common ground distributor. Two high-ampere cables (and a 12V cable) go to the top to be connected to the top plate to receive the input battery power.}
	\label{fig:CAD}
\end{figure}

The assembly of the cluster starts with the mechanical construction of the housing (Fig. \ref{fig:CAD}) and the installation of the electrical system of the bottom compartment. In parallel the middle compartment with the 16 nodes can be build (Fig. \ref{fig:boards}). The electrical components of the top plate can also be installed on it in parallel. 

Then the middle compartment is placed in the cluster housing. The power connectors of each node are fed through the hole of the middle compartment, as are the two power cables that go to the top plate. The 16 network cables that come from the nodes, the six cables from the top plate and an additional cable that will be connected to the synchronization node are all also fed through the hole to the bottom compartment and then connected to the industrial 24-port Gigabit Ethernet switch (OAM-600-65-4GX24GT). In order for the connection of the cable to the switch to be possible, the network cables have to be long enough to reach a little outside of the housing. A problem occurred after that: we used Cat6 network cables. These are quite stiff, so the switch could not be moved completely into the housing after connecting all 23 cables. We thus opted to have it stick out of the bottom compartment a bit, since this doesn't effect the operation of the cluster.

\begin{figure}[tbp]
	\centerline{
		\includegraphics[width=0.8\linewidth]{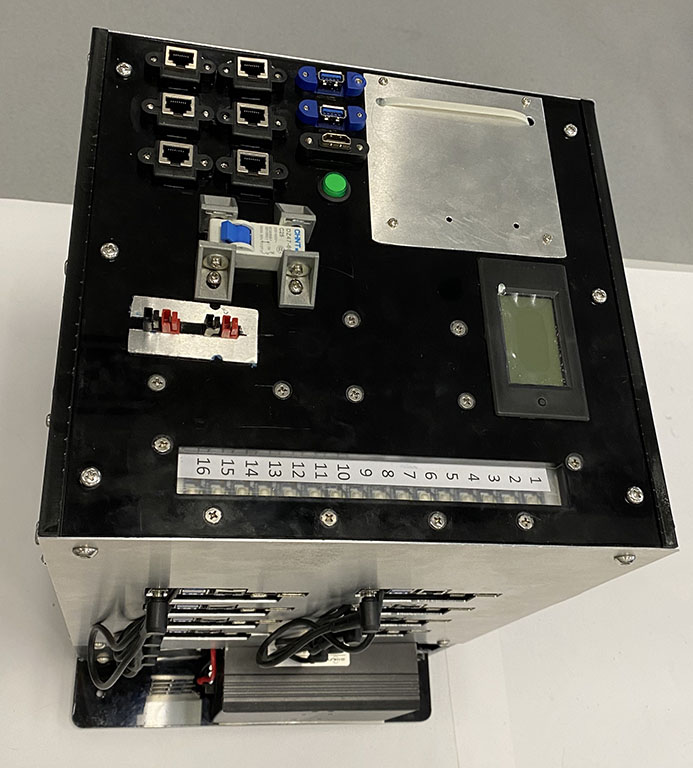}}
	\caption{The complete cluster. The top plate has several functions: For the power there are two pairs of power pole connectors, a circuit breaker and a display for the voltage, current, power and power used as well as power buttons and LEDs for all 16 nodes. Furthermore, there are six Ethernet RJ-45 sockets connected to the network switch. For the synchronization node that is to be added later there are already two USB sockets, a HDMI socket and a power button with LED as well as a 10cm $\times$ 10cm plate/ hole that will provide the connectors for the synchronization signals.}
	\label{fig:cluster}
\end{figure}

The two power cables from the bottom need to be connected to the top plate. The design of the holder for the 16 small PCBs with the power buttons allows us to first connect all cables to the PCBs and then mount the holder to the top plate.  The final cluster is depicted in Fig. \ref{fig:cluster}.

The CAD drawings, a part list as well as a video showing some of the assembly process are available\footnote{\url{https://robotics.shanghaitech.edu.cn/cluster_on_wheels}}.

Table \ref{tab:specs} shows a summary of the hardware specifications of the cluster.

\begin{table}[htbp]
	\caption{Cluster Specifications}
	\begin{center}
		\begin{tabular}{|c|c|}
			\hline
			\textbf{Item}& \textbf{Value} \\
			\hline 
			Dimensions Housing (w x d x h)&  288 x 270 x 400 mm\\ \hline
			Weight &  15kg \\ \hline
			DC Power In & 22 - 40 V \\ \hline

			Power Consumption ``Just Fans''/ Idle & 50/ 130 W \\ \hline
			Power Consumption Full Load  (Stress) &  750 W \\ \hline
			Node CPUs & 16 $\times$ AMD Ryzen 7 5700U \\ \hline
			Cores/ Threads & 128 / 256 \\ \hline
			RAM & 512GB \\ \hline
			Storage & 32TB  \\ \hline
			Storage write speed & 1.7GBps x 16  $\approx$ 27GBps \\ \hline
			Cost & approx. 25,000 USD \\ \hline
		\end{tabular}
		\label{tab:specs}
	\end{center}
\end{table}

\section{Cluster Software Stack}
\label{sec:software}

\begin{figure*}[htbp]
	\centerline{\includegraphics[scale=0.85]{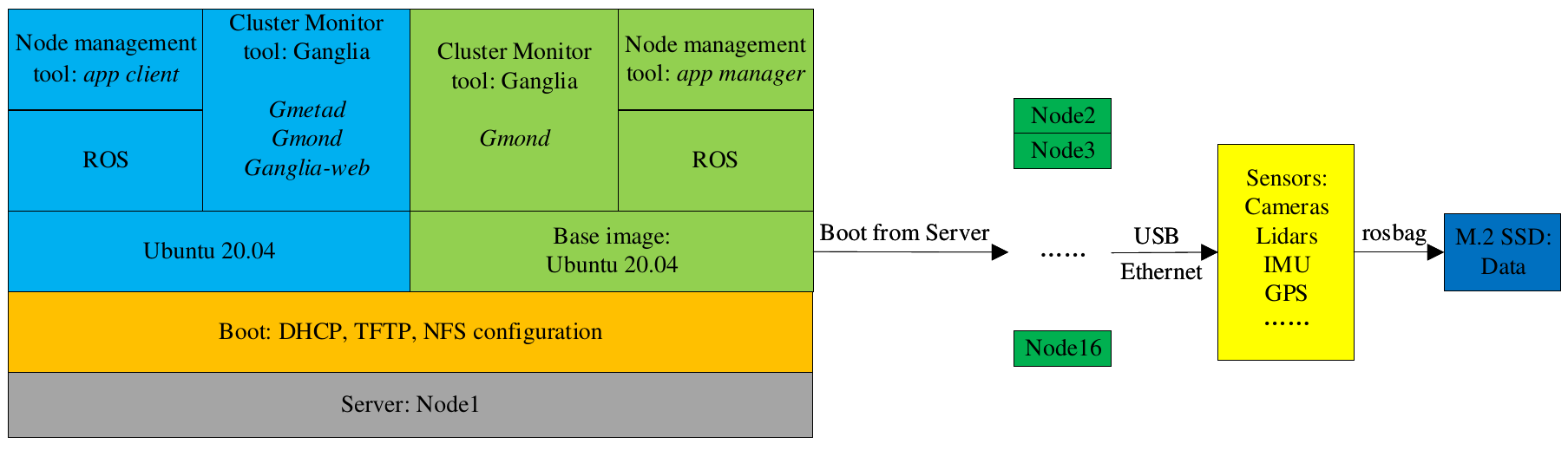}}
	\caption{The software stack of the Cluster for the ShanghaiTech Mapping Robot II. }
	\label{fig:structure}
\end{figure*}

For the cluster we chose to use Ubuntu 20.04 LTS as operating system with ROS Noetic Ninjemys as the robot operating system. Maintaining 16 of these installations is a hassle, so we opted to use network booting. For that we dedicate node 1 as the server and the 15 other nodes as clients. Fig. \ref{fig:structure} gives an overview of the software stack of our cluster. In the following software stack is described in detail.

%
%

\subsection{Boot on LAN}

Boot on LAN is a technique that lets a computer receive all files needed to start the operating system over the network instead of a local disk. This must be supported by the BIOS and the network card and properly configured in the BIOS. So we first must ensure that all client nodes have been set up to support the preboot execution environment (PXE), which is usually set on basic input output system (BIOS). The network hardware MAC addresses of all client nodes should be obtained. 

After installing Ubuntu 20.04 on the server (node 1) a number of services must be installed and configured. 

To build the cluster, it is necessary to use the dynamic host configuration protocol (DHCP) to assign IP addresses to nodes based on the MAC address that we previously noted.
Upon booting another node of the cluster, its BIOS will activate the network interface and receive its IP address via DHCP. It will then load the kernel of the cluster OS via TFTP. The trivial file transfer protocol (TFTP) is used for simple file transfers between a client and a server. After configuration of TFTP some necessary files need to  be prepared on the server for booting the client: kernel image, an initrd to go with it, a PXELINUX image, and a configuration file to go with the PXELINUX image. 

The initial kernel is transferred via TFTP, but the rest of the operating system needs a more sophisticated method. The initial kernel includes a NFS client, so we use NFS to provide all other operating system files. 

\subsection{NFS}
The network file system (NFS) is used to share a file system over the network. On the server {\it debootstrap} was used to create a basic Ubuntu 20.04 file system in the folder {\it /clusternfs}. This system is shared via NFS with the nodes, who mount it as their root file system. It is shared as "read only" to avoid various complications from multiple nodes writing to the same file system. Instead some crucial folders (/tmp, /var/log, /var/tmp) in the nodes are mounted as a temporary file system (tmpfs), for which the writes are saved in memory and are lost upon reboot.

\subsection{Client Setup}

All clients mount their SSD as the folder {\it /disk}. This is where, in the end, the collected sensor data will be stored. The {\it /disk} folder is shared via NFS by all the nodes and mounted by the server. This is mainly to be able to easily check the collected sensor data from the server. 

New software can be installed for the clients by logging into the server and there changing into the cluster file system on {\it /clusternfs} via the command {\it chroot}. Through this ROS and the node management tools described below are installed. 


\subsection{Nodes Management}
To comfortably operate the robot with its many sensors and ROS nodes, which are distributed over 16 cluster nodes, a graphical user interface (GUI) to start and stop the  ROS nodes is needed. For that we extended a node management tool called \textit{app manager} to work with multiple servers. It has two parts: \textit{app client} and \textit{app manager}. The \textit{app client} realizes an GUI which makes it easy to control and manage the ROS nodes of client nodes. The \textit{app manager} realizes start ROS nodes of client from master through LAN. 

\subsection{Cluster Monitoring}
In order to ensure that the system runs stably, a good way to monitor the many nodes is needed. The monitoring tool called \textit{Ganglia} can monitor a large number of nodes and only takes up very little CPU resources. It is mainly used to monitor system performance, such as: CPU, memory, hard disk utilization, I/O load, network traffic, etc. It is easy to see the working status of each node through the diagrams. The core of \textit{Ganglia} includes \textit{gmond}, \textit{gmetad} and a web front end: \textit{Ganglia-web}. \textit{Gmond} is run on the agent side, mainly used to collect the performance status of each node. \textit{Gmetad} is run on the server side and collects and stores the original data from \textit{gmond}. \textit{Ganglia-web} is a web service, which reads the data stored by \textit{gmetad} for web display.  

\subsection{Experiments}

To stress test the cluster system we utilized the Ubuntu program {\it stress}, which tests a computer by producing a high CPU, disk and memory load - similar to what we except when recording sensor data with the system. The power values from Table \ref{tab:specs} come from this test. 

\begin{figure}[tbp]
	\centerline{\includegraphics[scale=0.4]{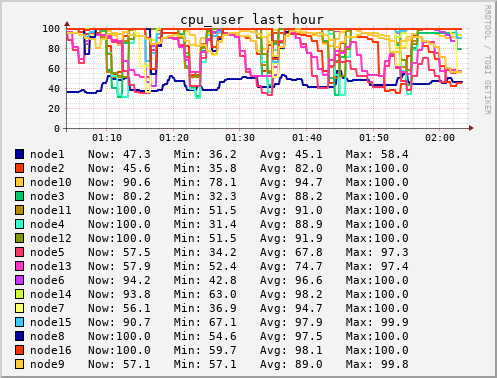}}
	\caption{One hour CPU load graph from the ganglia web interface from the two day stress test of the system.}
	\label{fig:ganglia_web}
\end{figure}

We also attempted a Linpack run of the cluster. One test ran for two days - Fig. \ref{fig:ganglia_web} shows a the CPU load graph of \textit{Ganglia} from that test. We utilized OpenMPI, AMD Optimizing CPU Libraries (AOCL) with a BLAS (Basic Linear Algebra Subprograms) component called BLIS and Linpack 2.3 for this. Unfortunately, due to our inexperience with Linpack, we were not able to find a configuration/ setup that resulted in meaningful results for the processing speed of the system.

%
%
%
%
%
%
%
%
%

\section{Conclusions}
\label{sec:conclusions}
In order to collect robotic datasets for SLAM research we are building a massive data collection robot. This robots needs a very big amount of I/O and processing power to collect, compress and store all the data in real time. After analyzing the requirements for the computing system of the robot and comparing several options, the actual solution, a 16-node cluster system, is presented. The hardware setup, including housing, DC power, networking, etc. is described and the software stack with booting from LAN and the app manager is presented. The working cluster was stress tested and found to be fully working.

Several improvements are planned on the cluster: {\it Power on LAN} is currently not functional - due to problems in the BIOS of the nodes. Hopefully this will be fixed with future BIOS updates. So currently nodes need to be started by hand using the buttons, as it is not an option to power on the nodes when they receive AC power, since the server node 1 must be up and running first. 

We may also opt to use more flimsy Ethernet cables in the cluster, in order to be able to fully insert the network switch to the housing. It may be possible to use unshielded cables, since the distances are very short. 

The next steps for us are the integration of the sensor synchronization node into the top plate and the actual construction of the ShanghaiTech Mapping Robot II.

\bibliographystyle{IEEEtran}

\bibliography{References}

\end{document}